\documentclass[10pt, conference, compsocconf]{IEEEtran}
\usepackage{amsmath,graphicx}
\usepackage{etoolbox}
\makeatletter
\patchcmd{\@makecaption}
  {\scshape}
  {}
  {}
  {}
\makeatother

\usepackage{fancyhdr} 
\usepackage{lipsum}   

\ifCLASSINFOpdf
\else
\fi
\fancypagestyle{firstpagefooter}{
  \fancyhf{}
  \fancyfoot[L]{\footnotesize This work has been published in [CIPCV 2024], DOI: [10.1109/CIPCV61763.2024.00025]. 
Copyright©[2024] IEEE. Personal use of this material is permitted. 
Permission from IEEE must be obtained for all other uses, in any current or future media, including reprinting/republishing this material for advertising or promotional purposes, 
creating new collective works, for resale or redistribution to servers or lists, or reuse of any copyrighted component of this work in other works.}
}

\IEEEoverridecommandlockouts

\usepackage{textcomp}
\usepackage{xcolor}
\def\BibTeX{{\rm B\kern-.05em{\sc i\kern-.025em b}\kern-.08em
    T\kern-.1667em\lower.7ex\hbox{E}\kern-.125emX}}

\usepackage{caption}
\begin{document}
\thispagestyle{firstpagefooter}
\lipsum[0]
\vspace*{10cm}
%
\title{A New Method For Optical Steel Rope Non-Destructive Damage Detection\\
\thanks{This research was funded by the Ministry of Science and Technology, the National Key Research and Development Program of China, grant number 2022YFC3005100.}}


\author{\IEEEauthorblockN{1\textsuperscript{st} Yunqing Bao}
\IEEEauthorblockA{\textit{Key Laboratory of Nondestructive Testing and} \\
\textit{Evaluation for State Market Regulation}\\
\textit{China Special Equipment Inspection and Research Institute}\\
Beijing, China \\
baoyunqing2@163.com}
\and
\IEEEauthorblockN{2\textsuperscript{nd} Bin Hu}
\IEEEauthorblockA{\textit{Key Laboratory of Nondestructive Testing and} \\
\textit{Evaluation for State Market Regulation}\\
\textit{China Special Equipment Inspection and Research Institute}\\
Beijing, China \\
hubin@csei.org.cn}
}


%


\maketitle

\begin{abstract}
This paper presents a novel algorithm for non-destructive damage detection for steel ropes in high-altitude environments (aerial ropeway). The algorithm comprises two key components: First, a segmentation model named RGBD-UNet is designed to accurately extract steel ropes from complex backgrounds. This model is equipped with the capability to process and combine color and depth information through the proposed CMA module. Second, a detection model named VovNetV3.5 is developed to differentiate between normal and abnormal steel ropes. It integrates the VovNet architecture with a DBB module to enhance performance. Besides, a background augmentation method is proposed to enhance the generalization ability of the segmentation model. Datasets containing images of steel ropes in different scenarios are created for the training and testing of both the segmentation and detection models. Experiments demonstrate a significant improvement over baseline models. On the proposed dataset, the highest accuracy achieved by the detection model reached 0.975, and the max F-measure achieved by the segmentation model reached 0.948.

\end{abstract}


%
\IEEEpeerreviewmaketitle

\section{Introduction}
In the realm of special equipment safety, such as amusement rides and elevators, the integrity of steel ropes plays a pivotal role \cite{xue2021research} \cite{doman2007amusement}. There are many different detection methods for steel wire ropes, including electromagnetic detection, ultrasonic guided wave, acoustic emission detection, etc. Optical detection, which is a method based on visual input, is an important way to judge the health condition of the steel ropes on the surface. Traditional optical inspection methods, primarily manual, not only struggle with detecting minute anomalies but also face significant challenges in accessing some steel ropes, particularly those suspended at high altitudes or in hard-to-reach locations, such as aerial ropeways. In this paper, we introduce a novel dual-model algorithm for steel rope damage detection in the open environment, leveraging the advancements in computer vision. The dual-model algorithm is composed of a segmentation model and a detection model.

The segmentation model, an innovative adaptation of the U-Net architecture, can process both RGB and depth images with the CMA(cross-modal attention) modules to segment steel rope objects effectively. The detection model, based on the small-object-friendly VovNet architecture, incorporates a DBB (Diverse Branch Block) \cite{ding2021diverse} module to get features with more diverse receptive fields and enhance detection capabilities. The detection model will judge whether a steel rope image is normal or abnormal. The comparison of our model (VovNetV3.5) with other state-of-the-art classification models is shown in Figure 1.

To conduct the experiments, we propose a dedicated steel rope dataset (about 700 images contained), tailored for training and testing both the detection and segmentation models.

We also propose a new background enhancement technique for the segmentation model with dual input. It improves the model's performance against unfamiliar natural backgrounds, which is a common challenge in real-world applications. 

The contributions of the paper can be summarized as follows:

1. Proposing a new U-Net architecture network that could segment the steel ropes with color and depth images as the input.

2. Designing a detection model in which the VovNetV2 is enhanced with DBB modules to improve performance in the task of anomaly detection on steel ropes. 

3. Introducing a steel rope dataset designed for segmentation and anomaly detection tasks, comprising approximately 700 steel rope images along with corresponding labels.

4. Introducing a new background enhancement method to improve the performance of the segmentation model in real scenarios.

\begin{figure}[t] 
  \centering
  \includegraphics[width= 9cm]{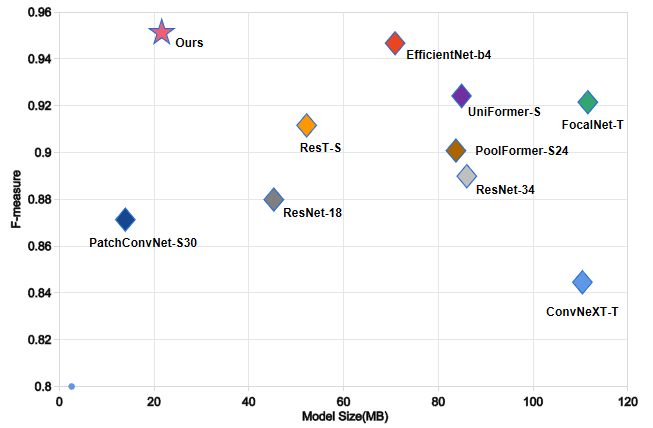} 
  \caption{Comparison of model size and performance of
 VovNetV3.5 with other state-of-the-art classification models. The F-measure is computed on our Rope-Seg dataset. The red star denotes our VovNetV3.5 model.}
\end{figure}

\section{Related Works}
\subsection{Optical Detection for Steel Rope}
The optical detection approach offers a highly effective and non-invasive method of inspection, particularly advanced in the field of Wire Rope (WR) surface damage assessment. This technique allows for a clear and immediate understanding of the condition of the steel wire rope's surface \cite{zhou2019review}.

For the task of automatic optical detection or segmentation of steel rope, many related studies are based on traditional machine learning or digital image processing methods. \cite{yaman2017auto}  presented a method for detecting anomalies in elevator ropes with conventional image processing algorithms and auto-correlation. The edge of the rope was extracted with the Sobel edge detection algorithm and the auto-correlation was applied to the edges to get the auto-correlation signal, which could be used to tell whether there is damage on the rope. \cite{sun2010texture} utilized a texture orientation detection method based on fuzzy Hough transform to determine the direction of the steel rope, and used an edge direction density histogram(EDDH) as the texture feature. Subsequently, the FCM clustering algorithm was employed to segment the texture of the steel rope image, and lines were detected from the clustering results to serve as the edge lines of the steel rope, taken as the segmentation result. In \cite{dong2013detection}, the images have been preprocessed using sharpening and equalization, followed by the application of the Canny operator to extract edges. Then the smoothness and entropy were calculated from the edges to distinguish normal steel ropes from damaged ones. \cite{liu2023wire} proposed a novel method, in which an Image Segmentation-based Central Multiscale Local Binary Pattern (ISCM-LBP) and Grey Level Co-occurrence Matrix (GLCM) feature fusion method is proposed. It performs PCA dimensionality reduction and GLCM feature fusion on features extracted from ISCM-LBP for a comprehensive image integrity description, and the defects are identified by a support vector machine (SVM) classifier. \cite{platzer2009hmm} proposed an algorithm where the steel ropes were analyzed using a Hidden Markov Model (HMM), with the histograms of oriented gradients (HOG) and their entropy taken as the features. And \cite{zhou2018health} applied a simple CNN architecture to detect various balancing tail ropes (BTR) faults, and the experiments show that the CNN architecture could perform better than k-nearest neighbor (KNN) and an artificial neural network with backpropagation (ANN-BP).

The aforementioned studies primarily focused on anomaly detection in steel ropes under relatively closed and idealized environments, where the background of the images is often uniform and unchanging, and the distance between the camera and the steel rope is relatively fixed and constant. However, in certain inspection scenarios, such as aerial ropeways, the background tends to be more complex, and the distance between the camera and the steel rope is relatively larger and constantly changing. Moreover, these studies mainly relied on traditional machine learning or image processing techniques (or very simple neural networks), lacking the comprehensive coverage and exploration of visual features in the scene by deep neural networks. Addressing these issues is the main goal of our research.

\subsection{Salient Object Detection}
Due to the relatively small area of steel ropes and the large area of backgrounds in the images, the algorithm often needs to segment the rope out before the detection. This kind of task aligns closely with salient object detection(SOD), which aims to identify and segment the most prominent parts of objects in images.

 Traditional methods in salient object detection primarily relied on hand-crafted features to create the saliency maps, such as \cite{srivatsa2015salient} and \cite{zhang2015minimum}. However, those approaches could not perform well in the complex and ever-changing environment. With the advent of deep learning after 2012, methods based on deep neural networks, especially CNN (Convolutional Neural Networks), have become predominant. 

Within the spectrum of deep learning methodologies, several employ purely convolutional frameworks. A notable example is the work presented in \cite{kim2016shape}, which detailed the development of a dual-branch CNN architecture adept at integrating both global and local information to accurately estimate the contours of the salient objects. Further, \cite{zhang2017learning} introduced an FCN (Fully Convolutional Network)-based model to effectively address the checkerboard artifact challenge using an innovative up-sampling process and the deep uncertain convolutional features (UCF). Many U-Struture models have good performance in the SOD task, the typical one is U$^2$Net \cite{qin2020u2}. The model's architecture is characterized by a nested U-structure. U$^2$Net is one of the best-performing methods in practical application scenarios nowadays. Other approaches like PiCANet, as discussed in \cite{liu2018picanet}, diverge from the conventional pure CNN framework. PiCANet's innovative strategy involves the development of a network that formulates attentive contextual features by selectively incorporating the features of surrounding regions. It is combined with the U-Net architecture to detect and segment salient pixel regions. PiCANet proves the significant role of the attention mechanism in SOD tasks.

In the paper, because of the limited number of training samples, the U-Net, which is small in size, is taken as the backbone of the segmentation model, and the attention structure named CMA is integrated to improve the performance. 

\section{Datasets}
\textbf{Rope-Seg dataset}
The dataset, which is named “Rope-Seg”, proposed for the segmentation model includes color images, corresponding depth images, and ground truth images (segmentation masks). The color and depth images are obtained using the RealSense and Zed2i cameras. All masks used as ground truth images are initially generated by the segment-anything model (SAM) and then manually adjusted. Additionally, some images, sourced from smartphones or the internet, do not have corresponding depth images. The dataset is divided into a training set and a test set. The training set comprises 556 groups of images, while the test set contains 157 groups of images. Most of the images are related to aerial ropeways, while a few of them are gathered from elevator scenarios or various scenes collected from the internet. We believe the diversity of the dataset could help to improve the model's performance in more generalized scenarios. Images in this dataset are shown in Figure 2.

\begin{figure}[htb]
\begin{minipage}[b]{1.0\linewidth}
  \centering
  \centerline{\includegraphics[width=8.5cm]{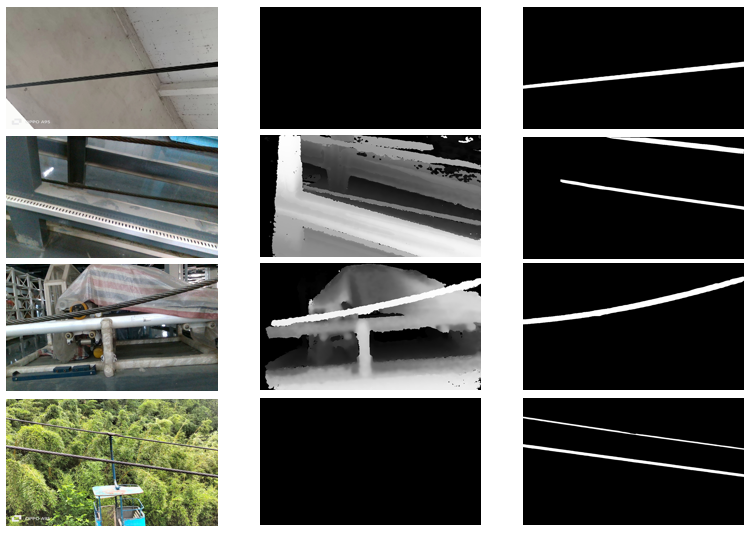}}
\end{minipage}
\caption{color image, depth image, and ground truth. A completely black image indicates there is no corresponding depth map. }
\end{figure}

\textbf{Rope-Detect dataset}
The dataset established for the detection model is similar to the structure of the segmentation model dataset, being divided into training and test sets. These sets contain both color images with corresponding depth images and color images without depth images. The segmentation masks for the steel ropes are manually extracted based on SAM, enabling the detection model to be directly trained and tested on these extracted steel ropes. The dataset is categorized into two groups: the images depicting abnormal steel ropes and the images depicting normal steel ropes. In the training set, there are 267 groups of abnormal images and 237 groups of normal images, totaling 504 images. In the test sets, there are 62 groups of abnormal images and 59 groups of normal images, making up 112 images in total. The dataset is named "Rope-Detect", and an illustrative comparison between normal and abnormal steel rope images is depicted in Figure 3.

\begin{figure}[htb]
\begin{minipage}[b]{1.0\linewidth}
  \centering
  \centerline{\includegraphics[width=8.5cm]{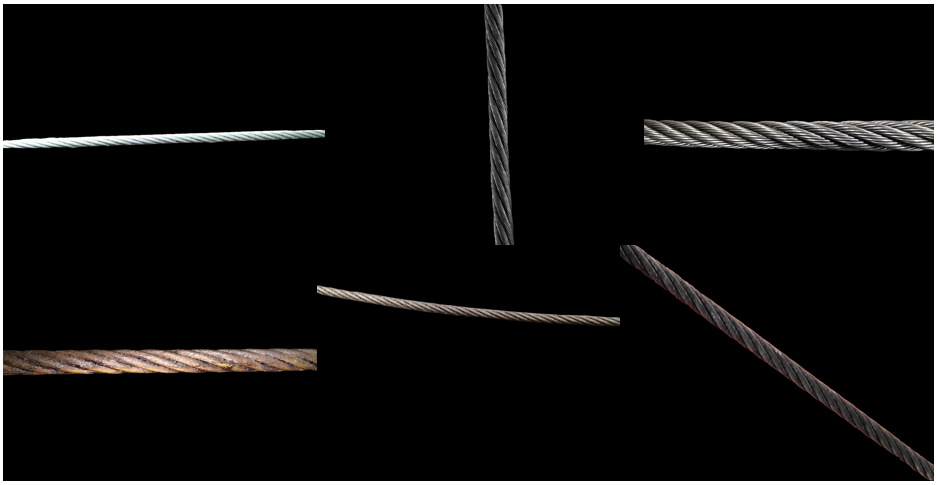}}
  \centerline{normal steel ropes}\medskip
\end{minipage}
\begin{minipage}[b]{1.0\linewidth}
  \centering
  \centerline{\includegraphics[width=8.5cm]{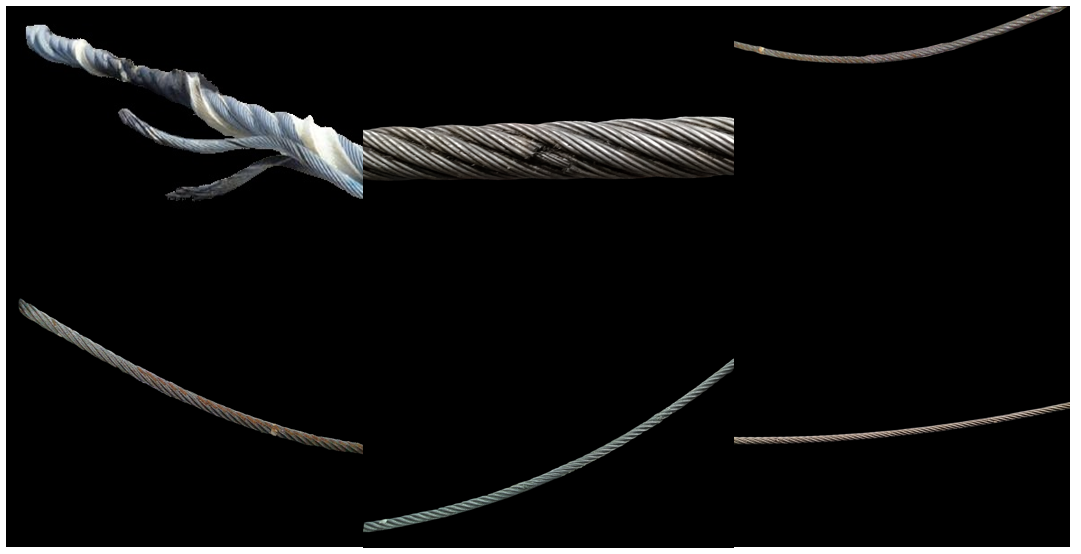}}
  \centerline{abnormal steel ropes}\medskip
\end{minipage}
\caption{Rope-Detect datasets}
\end{figure}

\section{Methodology}
\label{sec:pagestyle}
\subsection{Segmentation Model}
U-Net\cite{ronneberger2015u}, a convolutional neural network, is renowned for its outstanding performance in the domain of biomedical image segmentation. Its distinctive U-shaped architecture features a contracting path to capture image context and an expanding path for precise localization. This design allows U-Net to learn effectively from a limited number of training samples and achieve accurate pixel-level segmentation. 
 
In some special equipment scenarios such as amusement rides or elevator shafts, the target steel ropes and their background objects often have significant depth differences from the camera. This means that depth information can provide valuable guidance for the task of steel rope segmentation. Thus, an additional encoder pathway, mirroring the structure of the FuseNet \cite{hazirbas2017fusenet}, is incorporated into U-Net to fuse depth features with color features step by step, which refers to RGBD-UNet.

RGBD-UNet incorporates the Cross-Modal Attention (CMA) module. This integration is the key structure to fuse features extracted from color and depth images. Like FuseNet, the CMA modules are positioned before each pooling layer in the encoder part of U-Net to fuse the multi-modal data. As the early integration allows for a more nuanced and detailed feature representation, enhancing the model's segmentation capabilities. The detailed structure of the CMA module (shown in Figure 6) involves concatenating the feature maps which are extracted from color and depth input, followed by processing through spatial and channel attention modules. The pipeline of the segmentation model is shown in Figure 5.

\textbf{Background Augmentation.} 
The images of aerial ropeways, especially the ones with damage are quite hard to collect, and the backgrounds of the collected images are quite single. Thus we propose a background augmentation method to obtain the the images with varying backgrounds. For a color image and its corresponding depth image, the background augmentation algorithm randomly embeds the steel rope target into a natural background image which is taken from an actual scene, achieving a background replacement in the color image of the steel rope. Simultaneously, using image processing methods including morphological transformations, erosion, and dilation, the adjacent pixels of the steel rope in the original depth map are manipulated to generate the new depth map after the background replacement (we assume the distance between the steel rope and objects in the new background is no less than 3 meters, meaning it is beyond the range of depth measurement). The background augmentation method can help the model to perform robustly in the actual scenarios. The illustration of background augmentation is shown in Figure 4.

\begin{figure}[htb] 
  \centering
  \includegraphics[width= 8.5cm]{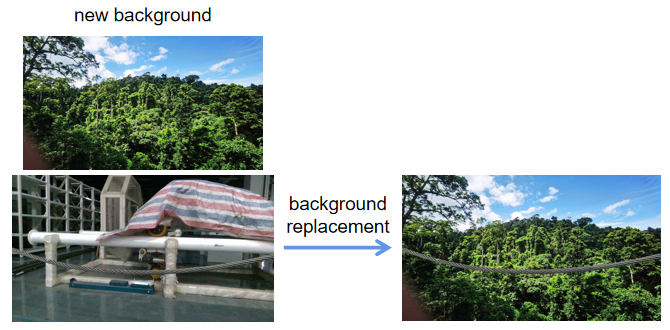} 
  \caption{Background augmentation process.}
\end{figure}

\begin{figure*}[t] 
  \centering
  \includegraphics[width=14cm]{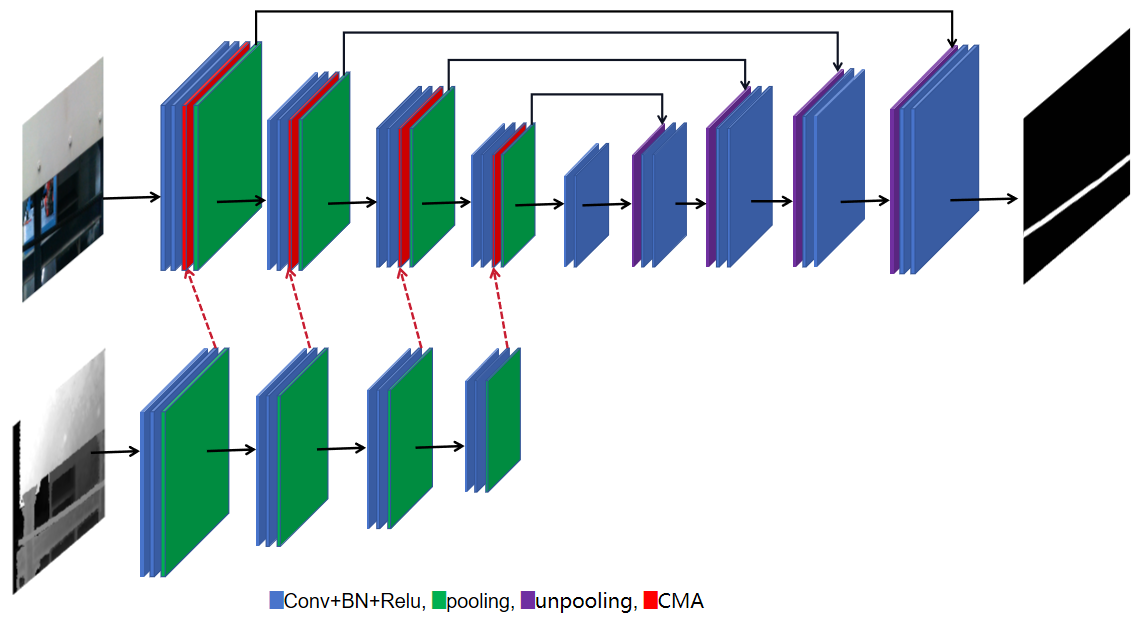} 
  \caption{The pipeline of the RGBD-UNet model}
\end{figure*}

\begin{figure}[t] 
  \centering
  \includegraphics[width= 8.5cm]{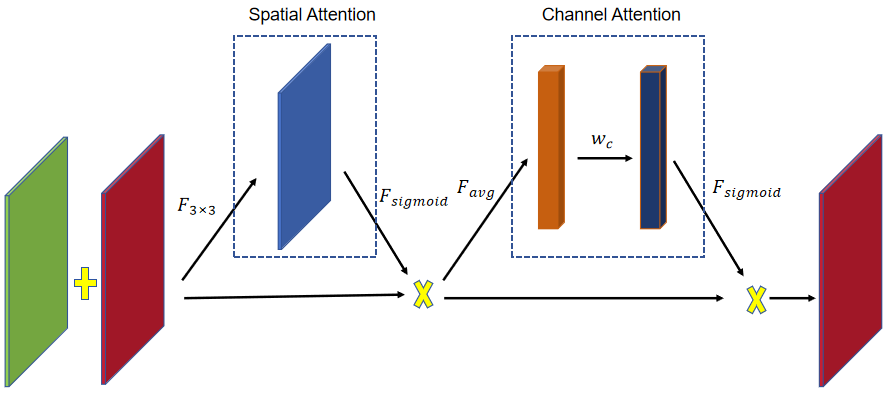} 
  \caption{The structure of the CMA module. Red denotes the feature maps extracted from the color input, and green denotes the feature maps extracted from the depth input. $F_{3x3}$ is a convolution layer that halves the number of channels, with kernel size 3. $F_{sigmoid}$ denotes the sigmoid function. $F_{avg}$ is the average pooling of the feature map.}
\end{figure}

\subsection{Detection Model}
To better detect defects on the segmented steel ropes, VovNetV3.5 is proposed based on VovNetV2 \cite{lee2020centermask} and VovNetV3 \cite{zhang2020acfd}. VovNet \cite{lee2019energy} is characterized by its One-Shot Aggregation (OSA) modules, which effectively capture diverse spatial features through densely connected layers. VovNetV2, based on the VovNet, incorporates the residual connections from ResNet \cite{he2016deep} and the eSE module inspired by SENet \cite{hu2018squeeze} to reduce optimization difficulty and enhance the representation ability of features. VovNetV2 has an exceptional capability in handling small objects, making it particularly suitable for detecting minor defects on steel ropes. These defects often occupy a very small proportion of the rope's surface area and are quite frequently seen on damaged ropes. Given the limited number of image samples in the dataset, VovNet-19 is adopted as the backbone network due to its minimal parameter count in the VovNet series, in case of overfitting.

To further improve the performance, the DBB \cite{ding2021diverse} module is integrated into the VovNetV2 model(shown in Figure 7), similar to how the ACB \cite{ding2019acnet} module is integrated into VovNet. The DBB module is developed from the ACB module. Different from ACB, the DBB module is a blend of convolution sequences, multi-scale convolutions, and average pooling rather than simple convolution kernels. This innovative design enriches the feature space with various scales and complexities, making it particularly effective for detecting subtle anomalies in steel ropes.

Furthermore, a single DBB or ACB module can be converted into a convolution layer through the re-parameterization process, which means no additional computational time during the inference stage. So we can enhance the model's performance without sacrificing inference speed. 

\begin{figure}[t] 
  \centering
  \includegraphics[width= 8.5cm]{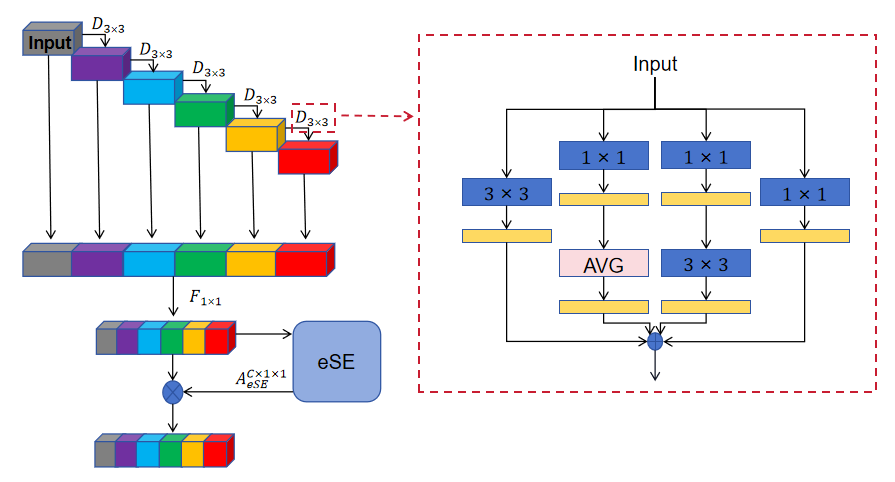} 
  \caption{The structure of VovNetV3.5, $D_{3x3}$ denotes DBB with kernel size 3. $F_{1x1}$ denotes the convolution layer with kernel size 1.}
\end{figure}

\section{Experiments}
\subsection{Segmentation Model}
We have conducted several experiments on the Rope-Seg dataset to prove the performance of the proposed segmentation method. All the experiments are completed on the GPU with a version of NVIDIA RTX A5000. The initial input size of images is 320*320, all batch sizes are set as 12, the initial learning rate is 5e-4, and the optimizer is AdamW. The datasets are augmented by horizontal flipping and random contrast. Also, all the images are randomly cropped from 320*320 to 288*288. The step learning rate scheduler is applied in the training stage. 

\subsubsection{Evaluation Metrics}
There are 5 metrics for evaluation: $F_\beta$ \cite{achanta2009frequency}, $E_m$ \cite{fan2018enhanced}, $MAE$ \cite{liu2019simple}, $S_m$ \cite{fan2017structure} and inference speed.

\textbf{The F-measure ($F_\beta$)},  originates from the PR (precision-recall) curve. Additionally, $MaxF_\beta$ represents the maximum value of $F_\beta$. This metric is capable of assessing both precision and recall simultaneously:

\begin{equation}
F_\beta = \frac{(1+\beta^2)\times Precision\times Recall}{\beta^2 \times Precision\times Recall}    
\end{equation}


\textbf{E-measure},  symbolized as $E_m$, integrates both low-level and high-level matching data, proving to be effective and efficient in its application. It can be calculated as follows:

\begin{equation}
E_m = \frac{1}{ w \times h}\sum_{r=1}^{H} \sum_{c=1}^{W}\phi_ {FM}\{r,c\}    
\end{equation}

$\phi_{FM}$ means the enhanced alignment matrix.

\textbf{MAE}, which stands for Mean Absolute Error, quantifies the pixel-by-pixel disparity between the predicted mask image and the actual ground truth mask image. $MAE$ can be computed as follows: 

\begin{equation}
MAE = \frac{1}{H\times W}\sum_{r=1}^{H} \sum_{c=1}^{W}|P(r,c) - G(r,c)|   
\end{equation}

P is the continuous prediction map and G is the ground truth map.

\textbf{S-measure ($S_m$)} represents structure-measure. This metric measures both the region-aware and the object-aware structural similarity. In the following formula, $S_o$ represents the object-aware structural similarity and $S_r$ represents the region-aware structural similarity.

\begin{equation}
S_m = \alpha \times S_o + (1-\alpha)\times S_r
\end{equation}

\textbf{Infer speed}, which is the abbreviation of inference speed, indicates how many images can be processed in one second.

\textbf{Model size}, the size of the model parameters saved from the PyTorch model.

\subsubsection{Results and Discussion}
The following experiments demonstrate the results of the ablation studies for different segmentation models based on the U-Net structure, which is conducted on the test set of the Rope-Seg dataset.

\begin{table*}[]
    \renewcommand\arraystretch{1}
    \centering
    \vspace{2mm}
    \normalsize
    \caption{The performances of different segmentation models. noCMA indicates directly adding depth features to color features without using CMA modules to fuse them. noaug and aug indicate whether the training set is augmented or not. For example, RGBD-UNet-aug refers to the RGBD-UNet model which has been trained on the augmented training set.}
    \begin{tabular}{|l|cccccccc|}
    \hline
     Model & Max$F_\beta\uparrow$ & Mean$F_\beta\uparrow$  & MAE$\downarrow$ & Max$E_m$ $\uparrow$ & Mean$E_m$ $\uparrow$ & $S_m$$\uparrow$ & Infer Speed $\uparrow$ & Model Size
        \\ \hline
        U-Net & 0.922 & 0.869 & 0.013 & 0.981 & 0.950 & 0.900 & 200FPS & 121.4MB
        \\ 
        U-Net-noaug & 0.861 & 0.803 & 0.024 & 0.943 & 0.906 & 0.863 & 200FPS & 121.4MB
        \\ 
        U-Net-aug & 0.908 & 0.849 & 0.018 & 0.967 & 0.929 & 0.887 & 200FPS & 121.4MB
        \\ 
        RGBD-UNet & 0.944 & 0.902 & 0.011 &0.986 & 0.965 & 0.927 & 139FPS & 166.6MB
        \\
        RGBD-UNet-noCMA & 0.932 & 0.888 & 0.012 &0.986 & 0.957 & 0.907 & 192FPS & 140.1MB
        \\
        RGBD-UNet-noaug & 0.889 & 0.832 & 0.019 &0.962 & 0.911 & 0.882 & 139FPS & 166.6MB
        \\
        RGBD-UNet-aug & 0.948 & 0.905 & 0.011 &0.985 & 0.964 & 0.934 & 139FPS & 166.6MB
        \\
        U$^2$Net & 0.934 & 0.881 & 0.011 &0.981 & 0.953 & 0.912 & 53FPS & 176.6MB
        \\ \hline
    \end{tabular}
\end{table*}

From Table 1, we can see that adding depth information effectively improves all metrics. Using the CMA module, instead of directly adding color and depth features, can further enhance performance, but with the cost of slightly lower inference speed. Figure 8 demonstrates how depth information contributed to generating superior saliency maps. Nevertheless, in unfamiliar backgrounds and when the depth information is unavailable, our segmentation model may produce unsatisfactory results, as exemplified in the last two rows of this Figure. After background augmentation, the training set expanded from 556 to 790 images, and the test set expanded from 157 to 223 images. Comparing the performance of U-Net-noaug with U-Net-aug, and RGBD-UNet-noaug with RGBD-UNet-aug, we find that augmenting the background of some images in the training set improves the model's performance on the augmented test set, where the backgrounds of some images are replaced. This suggests that background augmentation can enhance the model's generalization ability. Figure 9 shows the visual representation of the impact of background augmentation. Also, compared with u$^2$Net, our model can get slightly better results and achieve higher inference speed.

\begin{figure}[t] 
  \centering
  \includegraphics[width= 8.5cm]{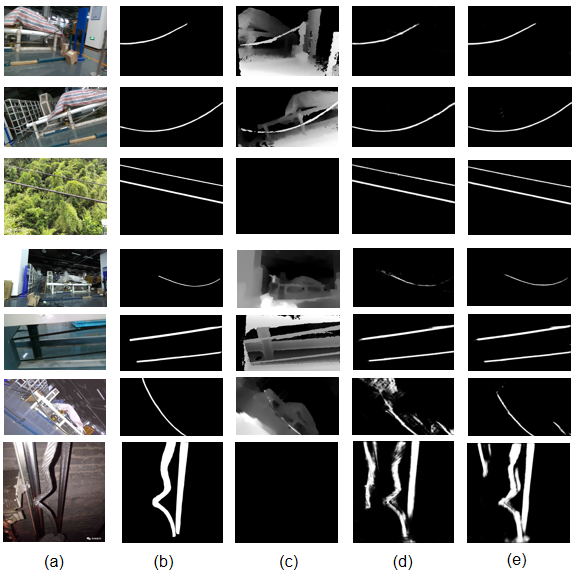} 
  \caption{Qualitative comparison of RGBD-UNet with U-Net. A completely black image indicates there is no corresponding depth input.(a) image, (b) ground truth, (c) depth image, (d) U-Net, (e) RGBD-UNet}
\end{figure}

\begin{figure}[t] 
  \centering
  \includegraphics[width= 8.5cm]{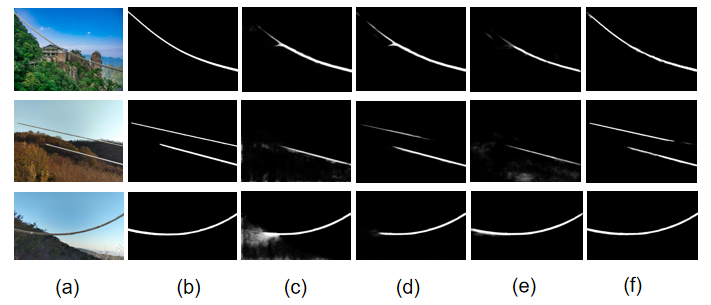} 
  \caption{Background augmentation method helps the models to get better performance in the unfamiliar background. (a) image, (b) ground truth, (c) U-Net-noaug, (d) U-Net-aug, (e) RGBD-UNet-noaug, (f) RGBD-UNet-aug}
\end{figure}

\begin{table*}[ht]
    \renewcommand\arraystretch{1}
    \centering
    \vspace{2mm}
    \normalsize
    \caption{Peformances of different versions of VovNet on Rope-Detect datasets.}
    \begin{tabular}{|l|c|c|c|c|c|c|}
    \hline
    Metric & VovNetV2  & VoVNetV3  & VovNetV3.5 & VovNetV2 & VovNetV3 & VovNetV3.5 \\ \hline
    Input Size & 224$\times$224 & 224$\times$224 & 224$\times$224 & 448$\times$448 & 448$\times$448 & 448$\times$448 \\  
     \hline
    Accuracy$\uparrow$ & 0.942 & 0.951 & 0.959 & 0.959 & 0.959 & 0.975 \\
    $F_\beta$ $\uparrow$ & 0.945 & 0.951 & 0.956 & 0.959 & 0.961 & 0.976 \\ 
    Model Size &10.4MB &19.3MB &21.2MB &10.4MB &19.3MB &21.2MB \\ \hline
    \end{tabular}
\end{table*}

\begin{table*}[ht]
    \renewcommand\arraystretch{1}
    \centering
    \vspace{2mm}
    \normalsize
    \caption{Results of different detection models. (*) indicates the metric is obtained from the test set generated by the RGBD-UNet model. }
    \begin{tabular}{|l|c|cccccc|}
    \hline
     Model & Input Size  & Accuracy$\uparrow$ & $F_\beta$ $\uparrow$  & Accuracy(*)$\uparrow$ & $F_\beta$(*)$\uparrow$ & Infer Speed $\uparrow$ &Model Size
        \\ \hline
        ResNet-18 \cite{he2016deep} & 224$\times$224 & 0.876 & 0.880 & 0.860 & 0.874 & 435FPS & 44.8MB
        \\
        ResNet-34 \cite{he2016deep} & 224$\times$224 & 0.884 & 0.891 & 0.868 & 0.877 & 270FPS & 85.3MB
        \\
        EfficientNet-b4 \cite{tan2019efficientnet} & 224$\times$224 & 0.942 & 0.944 & 0.917 & 0.919 & 63FPS & 71.0MB
        \\
        PatchConvNet-S30 \cite{touvron2021augmenting} & 224$\times$224 & 0.860 & 0.872 & 0.851 &0.859 & 67FPS& 14.0MB
        \\
        FocalNet-T \cite{yang2022focal} & 224$\times$224 & 0.917 & 0.921 & 0.884 & 0.892 & 90FPS & 111.6MB
        \\
        ResT-S \cite{zhang2021rest} & 224$\times$224 & 0.909
        & 0.912 & 0.868 & 0.873 & 100FPS & 52.8MB
        \\
        ConvNeXT-T \cite{liu2022convnet} & 224$\times$224 & 0.835 & 0.844 & 0.810 & 0.832 & 204FPS & 111.4MB
        \\
        UniFormer-S \cite{li2023uniformer} & 224$\times$224 & 0.926 & 0.927 & 0.909 & 0.911 & 125FPS & 84.3MB
        \\
        PoolFormer-S24\cite{yu2022metaformer} & 224$\times$224 & 0.901 & 0.903 & 0.868 & 0.877 & 134FPS & 83.6MB
        \\
        VovNetV3.5-19 & 224$\times$224 &  0.959 & 0.956 & 0.934 & 0.936 & 320FPS & 21.2MB
        \\
        VovNetV3.5-19 & 448$\times$448 &  0.975 & 0.976 & 0.942 & 0.945 & 320FPS & 21.2MB
        \\ \hline
    \end{tabular}
\end{table*}

\subsection{Detection Model}
We also conducted several experiments on the Rope-Detect dataset. The initial input size of images is scaled as 224*224 and 448*448. All batch sizes are set as 16, the initial learning rate is 5e-4, and the optimizer is AdamW. The datasets are augmented by horizontal flipping, random contrast, and random rotation. The cosine learning rate scheduler is applied in the training stage, with the learning rate oscillating between 5e-4 and 1e-6.  Additionally, all images input into the detection model are the result of foreground extraction based on either manually segmented masks or masks output by the segmentation model (RGBD-UNet).

\subsubsection{Evaluation Metrics}
There are 4 metrics for evaluation: Accuracy, $F_\beta$, inference speed, and the model size, which are all key metrics used in the evaluation of binary classification models.

\subsubsection{Results and Discussion}
Table 2 shows the performance of VovNetV2, VovNetV3, and VovNetV3.5 (all generated from the VovNet-19 structure) on the Rope-Detect dataset under different input size conditions, where the steel rope targets in the dataset are segmented from the image backgrounds based on manually annotated masks. The results reveal that VovNetV3.5 potentially outperforms VovNetV2 and VovNetV3. This performance improvement is more pronounced when employing a larger input size of 448x448, which contributes to an increase in all the metrics. Larger input sizes may lead to better performance because of the common occurrence of small areas worn on the surface of steel ropes. Besides, the improvement of the performance does not bring a decrease in inference speed, which is the advantage of the DBB module.


We compared our model with other state-of-the-art classification models, as shown in Table 3. The embedding dimension of PatchConvNet-S60 was halved in the number of layers in the column part to achieve better performance, then we get PatchConvNet-S30. All the models presented in the table are the best-performing versions selected. The results show that our model, VovNetV3.5-19 (based on VovNet-19), has the best overall performance and also a relatively high inference speed.

Also, We segmented the steel ropes from the test set of the Rope-Detect dataset using the RGBD-UNet model to form a new test set. Then we evaluated the performance of different classification models on this new test set, with the results presented in Table 3. It can be observed that due to the mask segmentation accuracy of the newly generated detection dataset not being as high as that of the manually segmented one, the performance of all detection models has declined to varying degrees. Our model still demonstrates the best performance, striking a balance between speed and accuracy.

\section{Conclusion}
In this paper, we propose a dataset for the segmentation and detection of the steel ropes used in amusement facilities and elevator-related scenarios. A new segmentation model named RGBD-UNet is proposed, which integrates the CMA module into the U-Net to effectively fuse color and depth information. We also propose a new detection model named VovNetV3.5, which combines the DBB module into VovNetV2. A series of experiments are conducted to prove the advantages of our models.
    
In future research endeavors, it might be beneficial to consider incorporating CMA modules and depth information into other U-structure models. Currently, due to time and human resources limitations, the classification of steel ropes is restricted to two broad categories: normal and abnormal. However, there is potential for future studies to delve into a more granular classification of abnormal steel ropes, distinguishing them into subcategories such as twisted, worn, frayed, and rust. Then it could yield more detailed insights into the condition of steel ropes.


\bibliographystyle{IEEEtran}
\bibliography{IEEEabrv,bare_conf}
\end{document}